\documentclass{article}
\usepackage[T1]{fontenc}
\usepackage{arxiv_preprint,times}
\usepackage{amsmath,amssymb,amsthm,mathtools}
\usepackage{graphicx,booktabs,algorithm,algpseudocode,float}
\usepackage[hidelinks]{hyperref}
\usepackage{xurl}
\hypersetup{pdftitle={Rethinking Successive Rejects Algorithms},pdfauthor={M. K. and J. K.}}
\title{Rethinking Successive Rejects Algorithms}
\author{Mohamed~Khair\\ Mohamed bin Zayed University \\ of Artificial Intelligence \\ \texttt{mohamed.khair@mbzuai.ac.ae} \And Junpei~Komiyama\\ Mohamed bin Zayed University \\ of Artificial Intelligence \\ RIKEN AIP \\ \texttt{junpei@komiyama.info}}

\newcommand{\R}{\mathbb R}
\newcommand{\Prob}{\mathbb P}
\newcommand{\E}{\mathbb E}
\newcommand{\bmP}{\boldsymbol P}
\newcommand{\bmQ}{\boldsymbol Q}
\newcommand{\PoE}{\mathrm{PoE}}
\newcommand{\Hone}{H_1}
\newcommand{\HSR}{\ensuremath{H_1}\textnormal{-SR}}
\newcommand{\barlog}{\overline{\log}}
\newcommand{\norm}[1]{\left\lVert#1\right\rVert}
\newcommand{\ind}{\mathbf 1}
\DeclareMathOperator{\KL}{KL}
\DeclareMathOperator*{\argmin}{arg\,min}
\DeclareMathOperator*{\argmax}{arg\,max}
\newtheorem{theorem}{Theorem}
\newtheorem{lemma}[theorem]{Lemma}
\newtheorem{proposition}[theorem]{Proposition}
\newtheorem{corollary}[theorem]{Corollary}
\theoremstyle{remark}

\preprintcopy

\begin{document}
\maketitle

\begin{abstract}
In the best-arm identification problem, an algorithm adaptively selects arms to find the arm of the largest mean reward. Successive Rejects (SR) is a simple fixed-budget best-arm identification algorithm that eliminates one candidate arm at each phase.
At first glance, SR seems to have limited adaptivity, since it follows predetermined phase lengths. However, for Gaussian arms with common variance, SR's worst-instance error exponent normalized by the popular complexity measure $H_1(P)$ is optimal up to a constant factor, which partly explains why SR's performance is competitive compared with state-of-the-art adaptive sampling algorithms. In this paper, we extend the framework of SR by optimizing its predefined schedule of phase lengths.
We formulate phase-schedule design as an optimization problem for the normalized error exponent.
Then, we introduce a version of SR with the optimized schedule, which we call $H_1$-balanced Successive Rejects (\HSR{}).
This schedule is optimal within the class of SR-type algorithms with deterministic equal-count schedules.
For Gaussian arms with common variance, $H_1$-SR achieves at least $11/12$
of the optimal normalized rate over all adaptive algorithms for every
$K\ge2$, and this approximation ratio converges to one as
$1-O(1/\log K)$ as $K\to\infty$. This is a significant improvement over original SR,
whose corresponding fraction of the optimal normalized rate converges to $1/4$.
Finally, experiments on ten synthetic instances and two dataset-derived Gaussian
benchmarks compare the optimized schedule with original SR at finite
horizons. They show substantial gains on many instances while also
exposing a regime in which more aggressive early elimination is costly.
\end{abstract}

\section{Introduction}

Best-arm identification asks how to use noisy measurements to select the
alternative with the largest mean. The forecaster, who follows an algorithm, adaptively chooses which
arm to measure and, after a prescribed budget of $T$ observations, outputs
a final recommendation: its estimate of the best arm. 

Successive Rejects (SR; \citealp{audibert2010best}) is a particularly simple
elimination algorithm that involves $K-1$ phases. Each phase gives the same number
of additional samples to every remaining arm, then eliminates the arm
with the smallest cumulative empirical mean. The phase lengths are fixed
in advance.
Sequential Halving (SH; \citealp{karnin2013almost}) uses a similar design
but eliminates approximately half the remaining arms in each phase.
These methods show that simple elimination can achieve an exponentially
small probability of error.
The multiplicative constant in the exponential rate, however,
determines how much budget is needed when small error probabilities matter.

The two methods also differ in how they use observations. Original SH
ranks arms using samples collected in the current phase, whereas SR
retains earlier samples in each surviving arm's empirical mean. This
distinction can affect finite-budget performance.
SR remains a competitive benchmark alongside more recent algorithms such as Continuous Rejects
\citep{wang2023large} and Almost Tracking \citep{komiyama2026anytime}.
Its continued relevance raises two questions: how close is its error
rate to the best attainable rate, and can a small change to its schedule
close the remaining gap?

The limited adaptivity of SR makes its performance gap particularly intriguing.
Once the rejection rule is fixed---sample all surviving arms equally and
eliminate an arm with the smallest cumulative empirical mean---the
only remaining design choice is the \emph{clock}: how many observations should each surviving arm have when each elimination takes
place. 
Original SR makes one particular choice. 
It is not immediate whether that choice is well matched to the statistical difficulty of
avoiding an erroneous elimination.

We therefore study the phase schedule itself as an optimization variable.
Our argument proceeds from the outside in. We first place an information
bound on the $H_1$-normalized rate of an arbitrary adaptive algorithm, where the $H_1$ is a popularly used complexity measure of the instance.
We then
restrict attention to the class of SR-type algorithms with deterministic equal-count
schedules, which is a subset of all adaptive algorithms that includes the original SR, and ask for the best schedule in that class. A sharp analysis of the event that the true best arm is simultaneously below all surviving
competitors turns the statistical problem into a small deterministic
balancing problem. Solving it produces the phase schedule; the algorithm
is a consequence of that optimization. 

\subsection{How should we compare identification algorithms?}

An algorithm can work well when one arm is clearly better and struggle
when several arms have nearly equal means. Comparing raw error
probabilities across such instances confounds algorithm quality with
problem difficulty. We therefore account for the gaps between arm means
before comparing how quickly errors decrease with the budget.
Following \citet{komiyama2022minimax,komiyama2026anytime}, we compare
the worst-instance exponential rate after normalization by an instance
complexity. We use $H_1$, the sum of inverse squared gaps to the best
arm, as a complexity measure that we normalize the error rate. 
A larger normalized rate means a stronger guarantee relative to
this measure of difficulty. 

Our benchmark is the best such rate over \emph{all} adaptive algorithms, not just elimination methods. The algorithm does not know the true gaps;
$H_1$ is used to evaluate it, not as an input. This criterion does not 
require matching a gap-informed oracle on every instance
\citep{degenne2023existence}. We give the formal definition in
Section~\ref{sec:model}.

Our algorithm can be used to the best-arm identification with sub-Gaussian rewards, including rewards supported in $[0,1]$ (or any finite region). 
Our optimality is for Gaussian rewards as we provides the error rate lower bound to that case. 

\subsection{Contributions}

\begin{figure}[t]
\centering
\includegraphics[width=\linewidth]{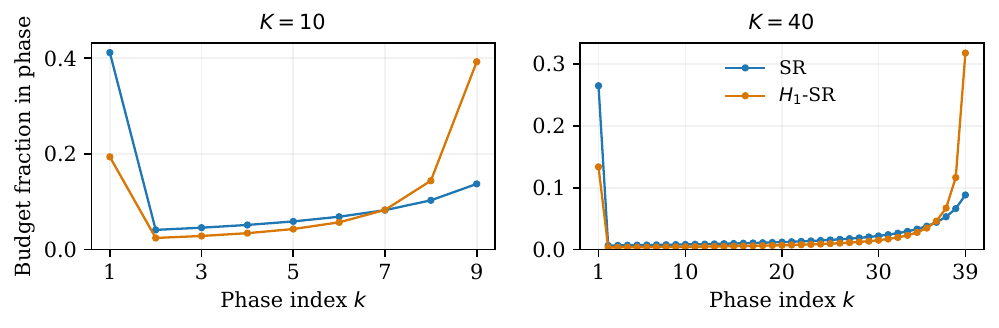}
\caption{Fraction of the total budget spent in each phase, for the original SR and our \HSR. 
Each phase includes all pulls of its surviving arms, not a cumulative
per-arm count. Both methods spend the same total budget and eliminate one arm per phase; \HSR{} spends more of the budget in the final phases. 
}
\label{fig:phases}
\end{figure}

The contributions follow the same progression as the analysis.
\begin{itemize}
\item \textbf{An explicit characterization of the best rate for any adaptive algorithm.} For Gaussian arms we derive an explicit upper bound $R_K^\star\le 1/(\sigma^2B_K)$ on the best
$H_1$-normalized rate attainable by arbitrary adaptive algorithms.
\item \textbf{Optimization within the class of SR-type algorithms.}
We characterize the worst-instance rate of deterministic equal-count successive-rejection schedules. Its unique solution gives us \HSR{} algorithm, that achieves the matching rate.
\item \textbf{Comparison with all algorithms and with original SR.} The
optimized schedule satisfies $R_{\HSR}/R_K^\star\ge B_K/D_K\ge11/12$ for
every $K$, with $B_K/D_K=1-O(1/\log K)$, whereas original SR satisfies $R_{\mathrm{SR}}/R_K^\star\to1/4$ as $K\to\infty$.
Figure \ref{fig:phases} illustrates the difference between SR and \HSR{} in terms of budget allocation across phases.
\item \textbf{Finite-budget experiments.} On ten synthetic and two
dataset-derived Gaussian benchmarks we compare $H_1$-SR with original SR.
The experiments show that $\HSR{}$ often improves terminal error, while a deliberately highlighted failure case shows why the asymptotic worst-instance guarantee should not be read as pointwise finite-budget dominance.
\end{itemize}

\subsection{Related work}\label{subsec_related}

\paragraph{Elimination algorithms}
Beyond SR \citep{audibert2010best} and Sequential Halving \citep{karnin2013almost,DBLP:conf/icml/ZhaoSSJ23}, \citet{shahrampour2017sequential} study
general sequential-elimination schedules, including nonlinear cumulative phase lengths. However, they have not characterized the optimal scheduling of the phase lengths.
\citet{wang2023large} develop a large-deviation analysis of
adaptive identification and introduce Continuous Rejects, which dynamically chooses the phase lengths.
\citet{barrier2023nonparametric} consider fixed-budget identification problems for a class of reward distributions bounded on $[0,1]$ and consider the SR algorithm therein.

\paragraph{Minimax rates and implementable algorithms.}
\citet{audibert2010best,carpentier2016tight} show that there is an $O(\log K)$ gap in the complexity of identifying the best arm between the fixed-budget setting (considered in this paper) and fixed-confidence settings, in which the algorithm can choose when to stop sampling.
A notable difference that subsequent research disclosed is that optimization in these two settings is fundamentally different.
Namely, the fixed-confidence setting allows a uniformly good algorithm. Namely, there exist several algorithms such as Track and Stop \citep{garivier2016optimal} and Top-two Thompson Sampling \citep{Russo2020,jourdan2023varepsilon,DBLP:conf/nips/Bandyopadhyay0A24} that achieve this property, meaning they perform optimally on all problem instances, whereas in the fixed-budget setting, no such uniformly good algorithm exists \citep{komiyama2022minimax,wang2023large,degenne2023existence,DBLP:conf/icml/WangAP24}. Therefore, the latter setting is closer to traditional statistical decision theory\footnote{Note that, in the statistical decision theory framework, it is usually called the risk measure.} \citep{Jenninson1982}, where the optimality depends on the choice of the complexity measure $H(P)$, such as admissibility \citep{imbens2025admissibility} and minimax optimality \citep{komiyama2022minimax}.
In particular, \citet{komiyama2022minimax} formulate minimax rates given a complexity measure $H(P)$, and introduce the delayed optimal tracking (DOT) algorithm. However, DOT is almost impossible to implement in practice as it needs to solve a dynamic program with a long horizon.
Later on, \cite{komiyama2026anytime} proposed a practical algorithm called simple tracking that does not require the information on the horizon $T$, and its version called almost tracking, which has a provable constant-factor rate guarantee. However, its constant on top of the rate guarantee is impractically large.

Our main results start from a more practical viewpoint. Starting from the popularly used SR algorithm, we improve its performance without introducing additional complexity into the algorithm.
Our main algorithm, \HSR{}, significantly improves the practical performance of SR, and we derive nearly minimax rate guarantees for the most popularly used complexity measure $H_1(P)$.

\section{From arbitrary adaptive algorithms to an optimized rejection schedule}
\label{sec:theory}

\subsection{Model and performance criterion}
\label{sec:model}

We follow the notation of \citet{komiyama2026anytime}. There are $K\ge2$
arms with unknown means $\bmP=(P_1,\ldots,P_K)$. At round $t$, algorithm
$\mathcal A$ chooses $I(t)\in[K]$ from the history up to round $t-1$,
then observes a fresh reward $X(t)$ from that arm. Write
\begin{equation}
 N_i(t)=\sum_{u=1}^t\ind\{I(u)=i\},\qquad
 Q_i(t)=\frac{\sum_{u=1}^t X(u)\ind\{I(u)=i\}}{N_i(t)}
 \quad(N_i(t)>0).
 \label{eq:observations}
\end{equation}
Thus $P_i$ is a true mean and $Q_i(t)$ is an empirical mean; write
$\bmQ(t)=(Q_1(t),\ldots,Q_K(t))$ once all arms have been sampled.
After the known horizon $T$, the algorithm recommends $J(T)$.

Assume a unique best arm and define
\[
 i^\star=i^\star(\bmP)=\argmax_{i\in[K]}P_i,\qquad
 P^\star=\max_iP_i,\qquad
 \Delta_i=P^\star-P_i>0\quad(i\ne i^\star).
\]
Parenthesized subscripts denote true ranks:
$P_{(1)}>P_{(2)}\ge\cdots\ge P_{(K)}$.
The unique-best assumption makes the target recommendation unambiguous
and ensures that every suboptimal gap is strictly positive.
For the exact-rate and minimax analysis, observations from arm $i$ are
independent draws from $N(P_i,\sigma^2)$, with a common variance
$\sigma^2>0$, and the reward streams are independent across arms.

We evaluate an algorithm through four quantities:

\begin{align}
 \PoE_{\mathcal A}(\bmP)&=\Prob_{\bmP}^{\mathcal A}
                  (J(T)\ne i^\star(\bmP)),&
 \Hone(\bmP)&=\sum_{i\ne i^\star(\bmP)}\Delta_i^{-2},\nonumber\\
 E_{\mathcal A}(\bmP)&=\liminf_{T\to\infty}
                  \frac1T\log\frac1{\PoE_{\mathcal A}(\bmP)},&
 R_{\mathcal A}&=\inf_{\bmP:\,\text{unique best}}
                  \Hone(\bmP)E_{\mathcal A}(\bmP).
 \label{eq:rate}
\end{align}
Here $\PoE_{\mathcal A}(\bmP)$ is the probability of recommending a
suboptimal arm after $T$ observations; as in
\citet{komiyama2022minimax}, its dependence on $T$ is implicit.
The complexity $H_1(\bmP)$ increases when competitors approach the best
arm, reflecting the greater difficulty of distinguishing their means.
The exponent $E_{\mathcal A}(\bmP)$ measures the asymptotic decrease of
error with budget: when a rate limit exists, the error behaves as
$\exp\{-T E_{\mathcal A}(\bmP)+o(T)\}$.
Finally, $R_{\mathcal A}$ is the smallest $H_1$-normalized exponent over
all unique-best instances. It measures the algorithm's worst-instance
guarantee after accounting for this complexity.

Let $\mathfrak A_K$ denote the class of all adaptive algorithm families
in this model, including randomized and horizon-dependent ones.
For each horizon $T$, arm selections may depend on the past observations
and internal randomness, and the final recommendation may depend on the
full observed history and internal randomness. These rules may depend
on $K$ and $T$, but not on the unknown means.
Our ideal objective is to attain the $H_1$-minimax rate
$R_K^\star=\sup_{\mathcal A\in\mathfrak A_K}R_{\mathcal A}$, or to guarantee a large
fraction of it with a practical algorithm. We determine SR's
normalized rate and improve it by changing only its phase lengths.

We exclusively consider $H_1$ here. Appendix~\ref{app:normalization} relates this convention to the other complexities studied in the literature \citep{audibert2010best,garivier2016optimal,komiyama2022minimax}.

For the finite-budget upper bound, we also allow independent rewards
with a common sub-Gaussian variance proxy $\sigma^2$: every centered
reward has moment generating function at most $e^{\lambda^2\sigma^2/2}$
for all $\lambda\in\R$. 
In particular, our algorithm also has error rate guarantees under any rewards supported in $[0,1]$ by setting $\sigma^2=1/4$ in the analysis.
For these extensions, the error probability depends on the full reward laws, and could be the case that we can further improve the bounds when we have additional knowledge about the reward distributions.

\subsection{An information bound for every adaptive algorithm}
\label{sec:all-algorithm}

We first describe the possible achievable rates for any adaptive algorithm, including not only SR-type algorithms,
but also Continuous Rejects \citep{wang2023large} and Simple Tracking \citep{komiyama2026anytime}. 
Namely, let
\[
 h_m=\sum_{j=1}^m j^{-1},\qquad
 B_K=\begin{cases}
 8,&K=2,\\
 43/7+2h_{K-1},&K\ge3.
 \end{cases}
\]
The constant $B_K$ is an inverse-rate information constraint.

\begin{theorem}[Rate upper bound]\label{thm:converse}
For Gaussian arms and the criterion in \eqref{eq:rate},
$R_K^\star\le 1/(\sigma^2 B_K)$.
\end{theorem}

The proof is deferred to Appendix~\ref{app:converse}, which gives the full
change-of-measure argument and derives the explicit constant $B_K$ from a
single limiting allocation constrained by a hierarchy of Gaussian alternatives.
The theorem is a worst-instance asymptotic restriction; it is not a
finite-budget lower bound for every individual instance.

\subsection{The class of SR-type algorithms}\label{sec:srclass}

We now focus on the class of SR-type algorithms.
We consider all deterministic schedules (ie., the phase lengths are fixed in advance) and we preserve SR's sampling and rejection rules. 
Let $\mathfrak A_K^{\mathrm{SR}}\subseteq\mathfrak A_K$ be the class of SR-type
algorithms that sample every surviving arm to the same predetermined
cumulative count, reject one arm with a smallest empirical mean, and
recommend the last survivor. With $s$ arms remaining, denote the
cumulative count by $m_s(T)$. The counts are positive integers satisfying
\[
 m_K(T)\le\cdots\le m_2(T),\qquad
 2m_2(T)+\sum_{s=3}^K m_s(T)\le T.
\]
The budget counts each rejected arm at its final sample count and both
finalists at $m_2(T)$. Tie breaking in elimination is fixed. 

In summary, an algorithm in $\mathfrak A_K^{\mathrm{SR}}$ is characterized by the choice of the deterministic schedule $(m_K(T),\ldots,m_2(T))$. For example, as we discussed in Section~\ref{sec:srclass}, the original version of SR \citep{audibert2010best} corresponds to $m_s/T \propto 1/s$. We next relate the deterministic schedule to the algorithm's performance (i.e., error rate $R_{\mathcal A}$).

The rest of this section shows how the deterministic schedule $(m_K(T),\ldots,m_2(T))$ can be optimized. 

\subsection{How the deterministic schedule controls performance}\label{sec:geometry}

Consider a phase with $s$ surviving arms, while the true best arm is
still active. Rejecting it requires its empirical mean to be no larger
than those of all $s-1$ competitors at once. To analyze this event
without conditioning on survival, fix a deterministic candidate set $A=\{i^\star,i_1,\ldots,i_{s-1}\}$ containing the true best.
Define its true gaps and complexity by
\[
 \delta_j=P_{i^\star}-P_{i_j}>0,\qquad
 H_A=\sum_{j=1}^{s-1}\delta_j^{-2},
\]
that is, $\delta_j = \Delta_{i_j}$.
For the proof, let $\overline Q_{i,m}$ be the mean of the first $m$
rewards in arm $i$'s potential independent sequence. This is an
equivalent way to construct the sequential observations, not an extra
observation available to the algorithm. At any deterministic common count $m$, the empirical means of the arms in $A$ are represented by $\overline Q_{i,m}$.

We now relate the deterministic schedule to the SR-type algorithm's error exponent
through concentration inequalities. 
Namely, for Gaussian rewards with variance $\sigma^2$, the probability
that an arm's empirical mean falls below a fixed threshold
$q<P_i$ satisfies
\begin{equation}
 \Prob\!\left(\overline Q_{i,m}\le q\right)
 \le \exp\!\left(
 -m\KL\!\left(N(q,\sigma^2)\,\|\,N(P_i,\sigma^2)\right)
 \right)
 = \exp\!\left(-\frac{m(q-P_i)^2}{2\sigma^2}\right).
 \label{eq:largedeviation}
\end{equation}

After subtracting $P^\star$, the true mean vector on $A$ is
$p_A=(0,-\delta_1,\ldots,-\delta_{s-1})$.
A necessary condition for rejecting the best arm is that the shifted
empirical vector belongs to
$\mathcal C_A=\{q\in\R^s:q_0\le q_j,\ 1\le j<s\}$.
This is a closed convex polyhedral cone, the intersection of the
half-spaces $q_0-q_j\le0$. It contains exactly the vectors in which the
true best arm has a smallest empirical mean, allowing ties. 
The quantity $d_A$ is the smallest Euclidean displacement of the
\emph{whole mean vector} that makes the true best arm empirically worst:
\begin{equation}
 d_A^2=\inf_{q\in\mathcal C_A}\|q-p_A\|^2
 =\min_{u\ge0}\left\{u^2+
             \sum_{j=1}^{s-1}(\delta_j-u)_+^2\right\}
 \ge\frac{(s-1)^2}{sH_A}.
 \label{eq:geometry}
\end{equation}
Here $(x)_+=\max\{x,0\}$ and $u$ is the downward displacement of
the best-arm empirical mean from $0$ to $-u$. For this fixed coordinate,
the nearest feasible competitor mean is $\max\{-\delta_j,-u\}$,
requiring an upward displacement $(\delta_j-u)_+$. Summing these squared
displacements and minimizing over $u$ gives the middle expression.
The inequality is tight when the gaps in $A$ are equal. Projection onto
this cone and a sub-Gaussian Chernoff bound, which gives a finite-sample
guarantee with the same quadratic structure in the exponent as
\eqref{eq:largedeviation}, give
\begin{equation}
 \Prob\!\left(\overline Q_{i^\star,m}\le
       \overline Q_{i,m}\ \text{for every }i\in A\setminus\{i^\star\}\right)
 \le \exp\left(-\frac{m(s-1)^2}{2s\sigma^2 H_A}\right).
 \label{eq:joint}
\end{equation}
Appendix~\ref{app:gaussian} proves both the projection step and the sharp
inverse-gap inequality. 
The large-deviation analysis of SR in
\citep{wang2023large,komiyama2026anytime} identifies the cost of such
misleading empirical means. Here we bound that cost, at the end of a phase $K+1-s$ where $s$ arms remains, sharply in terms of
$H_A$. Since $H_A \le H_1$, \eqref{eq:joint} also yields an upper bound in terms of $H_1$ that is uniform over all such sets $A$.
The analysis is enabled by the fact that an SR-type algorithm conducts uniform exploration at each phase, which is much more simpler than the design of algorithms appearing in \citep{wang2023large,komiyama2026anytime}.

The bound \eqref{eq:joint} is the key design input. Its coefficient depends only on the number $s$ of surviving arms. The next subsection packages this coefficient into a convenient quantity and asks how to choose all checkpoint counts so that no active-set size becomes the bottleneck.

\subsection{A small balancing problem with an explicit solution}\label{sec:balance}

The simultaneous-rejection bound in Section~\ref{sec:geometry} identifies
\[
 a_s=\frac{2s}{(s-1)^2},\qquad 2\le s\le K,
\]
as the information requirement associated with the checkpoint having $s$
surviving arms. Recall that $m_s(T)$ is the cumulative count per surviving arm
at the checkpoint with $s$ arms. 
An SR-type algorithm is characterized by $(m_s(T))_{2\le s\le K}$. Since this is a vector sequence of $T$, we take the limit as $T\to\infty$.
When the limits exist,
\[
 q_s=\lim_{T\to\infty}\frac{m_s(T)}{T},\qquad 2\le s\le K.
\]
Thus $q_s$ is a cumulative per-arm fraction, not the fraction spent in
that phase. 
For schedules without full limits,\footnote{Of course, such cases, for example, $m_s(T)$ quickly switches for different $T$, do not improve the rate!} Appendix~\ref{app:class} uses the same notation for limits along a common subsequence.
Since the error exponent is determined by the worst phase, the guaranteed normalized exponent is $\min_s q_s/(\sigma^2a_s)$.
Maximizing this among the class of limit fractions $(q_2,\ldots,q_K)$ gives
\begin{equation}
 \max_{\substack{0\le q_K\le\cdots\le q_2\\
                 2q_2+\sum_{s=3}^Kq_s\le1}}
       \min_{2\le s\le K}\frac{q_s}{a_s}.
 \label{eq:balance}
\end{equation}
Define the normalizing constant
\begin{equation}
 D_K:=2a_2+\sum_{s=3}^K a_s
 =4+2h_{K-1}+2h_{K-1}^{(2)},\qquad
 h_m^{(2)}=\sum_{j=1}^m j^{-2}.
 \label{eq:constants}
\end{equation}
The optimal limit fractions are given by $q_s$ such that \eqref{eq:balance} holds, is achieved by the following:
\begin{proposition}[Information balancing]\label{prop:balance}
The value of \eqref{eq:balance} is $1/D_K$, and its unique optimizer is
$q_s=a_s/D_K$ for all $s$.
\end{proposition}
\begin{proof}
If the minimum is at least $v$, every $q_s\ge va_s$. The budget gives
$1\ge v(2a_2+\sum_{s=3}^Ka_s)=vD_K$. Equality is feasible because
$a_s$ decreases with $s$, and equality in the budget forces equality
in every coordinate lower bound.
\end{proof}

\subsection{The resulting algorithm and its guarantee}\label{sec:algorithm}

Proposition~\ref{prop:balance} determines the asymptotically balanced checkpoint fractions. We implement those fractions at finite horizon by integer rounding. For $T\ge K$, let $A_1=[K]$ and $n_0=0$. In phase
$k\in\{1,\ldots,K-1\}$, the active set has size $s=K+1-k$. Set
\begin{equation}
 n_k=1+\left\lfloor\frac{T-K}{D_K}\,a_{K+1-k}\right\rfloor.
 \label{eq:clock}
\end{equation}
The checkpoint $n_k$ is the cumulative sample count \emph{per surviving
arm}, not the global time. Each arm in $A_k$ receives $n_k-n_{k-1}$
additional samples. At the resulting global round $t_k$, all active
arms satisfy $N_i(t_k)=n_k$, and an arm minimizing the empirical mean $Q_i(t_k)$ is
removed. The phase therefore uses $(K+1-k)(n_k-n_{k-1})$ pulls.
Algorithm~\ref{alg:hsr} summarizes the procedure.

The following algorithm bounds the probability of error for Algorithm~\ref{alg:hsr}.
\begin{algorithm}[t]
\caption{$H_1$-balanced Successive Rejects (\HSR)}
\label{alg:hsr}
\begin{algorithmic}[1]
\Require Number of arms $K\ge2$; horizon $T\ge K$; fixed tie-breaking order
\State $A_1\gets[K]$, $n_0\gets0$; compute $D_K$ from \eqref{eq:constants}
\For{$k=1,\ldots,K-1$}
  \State $s\gets K+1-k$; $n_k\gets1+\lfloor(T-K)a_s/D_K\rfloor$
  \State Pull each $i\in A_k$ exactly $n_k-n_{k-1}$ additional times
  \State Let $t_k$ be the current round; $j_k\gets\argmin_{i\in A_k}Q_i(t_k)$
  \State $A_{k+1}\gets A_k\setminus\{j_k\}$
\EndFor
\State Set $J(T)$ to the unique element of $A_K$
\State If fewer than $T$ pulls were used, pad the budget and ignore those observations
\State \Return $J(T)$
\end{algorithmic}
\end{algorithm}

\begin{theorem}[Finite-budget error upper bound]\label{thm:balanced}
For every $T\ge K$ and every unique-best instance with independent
$\sigma^2$-sub-Gaussian arm rewards,
\begin{equation}
 \PoE_{\HSR}(\bmP)\le
 \min\left\{1,(2^{K-1}-1)
 \exp\left(-\frac{T-K}{\sigma^2D_K\Hone(\bmP)}\right)\right\}.
 \label{eq:finite}
\end{equation}
\end{theorem}
Consequently, at every fixed instance the error exponent is at least
$1/(\sigma^2D_KH_1(\bmP))$. This gives
$R_{\HSR}\ge1/(\sigma^2D_K)$. 

In the Gaussian model, the asymptotic rate guarantee is tight:
Theorem~\ref{thm:class} below provides the matching upper bound,
yielding $R_{\HSR}=1/(\sigma^2D_K)$.

\subsection{Class optimality and comparison with any algorithm}\label{sec:classopt}

Since \HSR{} is designed to optimize the allocation of pulls across arms in a successive rejects framework, it achieves the best possible error exponent within this class of algorithms.
\begin{theorem}[Optimal clock within the class of SR-type algorithms]\label{thm:class}
For Gaussian arms,
$\sup_{\mathcal A\in\mathfrak A_K^{\mathrm{SR}}}R_{\mathcal A}=1/(\sigma^2D_K)$,
attained by \HSR.
\end{theorem}

The upper bound is enforced by a different hard instance at each
checkpoint. Consider $s$ arms with small gaps and $K-s$ arms with much
lower means. With high probability the latter are removed first, leaving
the difficult comparison to the $s$-arm checkpoint. Too few samples at
that checkpoint make best-arm rejection too likely. Applying this
argument to every $s$ yields simultaneous lower requirements on the
counts; their sum exhausts the budget at the proposed schedule.
Appendix~\ref{app:class} makes this argument precise.

This theorem identifies the best rate within $\mathfrak A_K^{\mathrm{SR}}$. Combining
it with the bound for arbitrary adaptive algorithms gives a separate
comparison with the unrestricted optimum.

\begin{corollary}[Uniform approximation and leading constant]\label{cor:factor}
For Gaussian arms and every fixed $K\ge2$,
\begin{equation}
 \frac{1}{\sigma^2D_K}\le R_K^\star\le\frac{1}{\sigma^2B_K},
 \qquad
 \frac{R_{\HSR}}{R_K^\star}\ge\frac{B_K}{D_K}
 \ge\frac{11}{12}.
 \label{eq:ratio}
\end{equation}
Moreover $B_K/D_K=1-O(1/\log K)$. As $K\to\infty$ after the
fixed-$K$ horizon limit, $R_{\HSR}/R_K^\star\to1$ and
$R_K^\star\sim1/(2\sigma^2\log K)$.
\end{corollary}
Corollary \ref{cor:factor} shows that \HSR{} is very close to the optimal rate attainable by any algorithm, which is somewhat surprising given the fact that it determines the length of the schedule beforehand and does not revisit the arms once they are eliminated.

\section{$H_1$-SR versus original Successive Rejects}\label{sec:srcompare}

\subsection{Sharp normalized-rate comparison}

For $T>K$, original SR uses
\begin{equation}
 n_k^{\mathrm{SR}}=
 \left\lceil\frac{T-K}{\barlog(K)(K+1-k)}\right\rceil,
 \qquad \barlog(K)=\frac12+\sum_{i=2}^K\frac1i.
 \label{eq:srclock}
\end{equation}

\begin{theorem}[Sharp normalized original-SR comparison]\label{thm:sr}
For Gaussian arms under \eqref{eq:rate},
$R_{\mathrm{SR}}=1/(8\sigma^2\barlog(K))$.
For $T>K$,
\[
 \PoE_{\mathrm{SR}}(\bmP)\le
 \min\left\{1,\frac{K(K-1)}{4}
       \exp\left(-\frac{T-K}{8\sigma^2\barlog(K)\Hone(\bmP)}\right)\right\}.
\]
Consequently,
\[
 \frac{R_{\mathrm{SR}}}{R_K^\star}
 \ge\frac{B_K}{8\barlog(K)}\ge\frac14.
\]
\begin{equation}
 \frac{R_{\HSR}}{R_{\mathrm{SR}}}
     =\frac{8\barlog(K)}{D_K}\longrightarrow4,
 \qquad
 \frac{R_{\mathrm{SR}}}{R_K^\star}\longrightarrow\frac14.
 \label{eq:srgain}
\end{equation}
\end{theorem}
Theorem \ref{thm:sr} shows that, for $K\ge3$, \HSR{} achieves a strictly better worst-instance normalized exponent compared to the original SR algorithm, and it is by $4$ for large $K$.

For general $\sigma^2$-sub-Gaussian arms, the displayed finite-budget
bound remains valid with prefactor $K(K-1)/2$ instead of $K(K-1)/4$.
The smaller Gaussian prefactor uses an additional one-half tail factor;
the exact-rate and minimax comparisons in the theorem are Gaussian.

The exact error exponent of original SR is established in prior work \citep{DBLP:conf/icml/WangAP24,komiyama2026anytime}. 
Appendix~\ref{app:sr} gives the short
self-contained argument needed for the $H_1$ normalization used here.

\subsection{Intuition on the ratio}

Recall that $m_s(T)$ is the cumulative number of samples per
surviving arm at the phase with $s$ arms remaining.
Ignoring integer rounding, the two schedules satisfy
\[
\text{Original SR:}\quad
m_s(T)\approx\frac{T}{s\barlog(K)},
\qquad
\text{\HSR{}:}\quad
m_s(T)\approx\frac{T}{D_K}\frac{2s}{(s-1)^2}.
\]
Their ratio is therefore
\[
\frac{m_s^{\HSR}(T)}{m_s^{\mathrm{SR}}(T)}
\longrightarrow
\frac{2\barlog(K)}{D_K}
\left(\frac{s}{s-1}\right)^2
\qquad (T\to\infty).
\]
The ratio increases as fewer arms remain: $H_1$-SR spends less before early
eliminations and more on the final comparisons (Figure~\ref{fig:phases}). Since
$2\barlog(K)/D_K\to1$, the final-two ratio tends to $4$, matching the
large-$K$ rate improvement in Theorem~\ref{thm:sr}.

\section{Numerical Experiments}
\label{sec:experiments}

The theory compares worst-instance normalized asymptotic exponents and does
not imply pointwise finite-budget dominance. We therefore compare $H_1$-SR
with original SR on twelve fixed-horizon Gaussian problems: ten synthetic
instances with $K=40$, a MovieLens 1M-derived instance with $K=31$, and an
Open Bandit-derived instance with $K=80$. The synthetic configurations follow \citet{komiyama2026anytime} and are
listed in Appendix~\ref{app:experiment-details}; the other two instances use
means derived from MovieLens 1M \citep{movielens1m} and Open Bandit \citep{SaitoOBP}. Rewards are unit-variance
Gaussian. Each cell uses $10{,}000$ repetitions at the configured horizon $T$.

Our primary empirical quantity is the ratio
$\widehat{\PoE}_{H_1\text{-SR}}/\widehat{\PoE}_{\mathrm{SR}}$: values below
one favor $H_1$-SR and values above one favor SR. Paired terminal comparisons
use two-sided exact McNemar tests with Holm correction over the twelve
predeclared comparisons at family-wise level $0.05$.

\begin{table}[H]
\centering
\small
\setlength{\tabcolsep}{3.8pt}
\renewcommand{\arraystretch}{0.98}
\caption{Terminal error (\%) over $10{,}000$ repetitions. Ratio $=\widehat{\PoE}_{H_1\text{-SR}}/\widehat{\PoE}_{\mathrm{SR}}$; boldface marks a Holm-corrected supported improvement.}
\label{tab:experiments}
\begin{tabular}{lrrrrrc}
\toprule
& & & \multicolumn{2}{c}{Terminal $\widehat{\PoE}$ (\%)} & & \\
\cmidrule(lr){4-5}
Instance & $K$ & $T$ & $H_1$-SR & SR & Ratio & Supported \\
\midrule
Synthetic 1  &40& 6,000 & \textbf{9.92} &18.05&0.550& $H_1$-SR\\
Synthetic 2  &40& 1,000 & \textbf{1.37} & 4.61&0.297& $H_1$-SR\\
Synthetic 3  &40& 4,000 & 7.72 & 7.45&1.036& --\\
Synthetic 4  &40& 4,500 & \textbf{9.88} &14.56&0.679& $H_1$-SR\\
Synthetic 5  &40&25,000 & \textbf{9.47} &13.73&0.690& $H_1$-SR\\
Synthetic 6  &40& 5,000 & \textbf{5.54} &11.30&0.490& $H_1$-SR\\
Synthetic 7  &40& 8,000 &15.88 & \textbf{4.98}&3.189& SR\\
Synthetic 8  &40& 2,000 & \textbf{2.01} & 4.17&0.482& $H_1$-SR\\
Synthetic 9  &40& 1,500 & \textbf{1.76} & 5.08&0.346& $H_1$-SR\\
Synthetic 10 &40& 2,500 & \textbf{7.03} &12.36&0.569& $H_1$-SR\\
MovieLens 1M &31&10,000 &\textbf{18.27}&24.87&0.735& $H_1$-SR\\
Open Bandit   &80& 3,000 & 9.37 &10.27&0.912& --\\
\bottomrule
\end{tabular}
\end{table}

$H_1$-SR has lower empirical terminal error on ten of twelve instances; after
Holm correction, nine comparisons support $H_1$-SR, one supports SR, and two
are inconclusive. MovieLens has ratio $0.735$, whereas synthetic instance 7
has ratio $3.189$. In that instance, $H_1$-SR uses $27$ observations per arm
before the first elimination versus $53$ for SR, but $2023$ versus $1054$ at
the final-two checkpoint. This illustrates the finite-budget early/late
allocation tradeoff. Appendix~\ref{app:experiment-details} visualizes all trajectories.
Regarding the synthetic instance 7, Appendix~\ref{app:non-dominance} defines dominance and discusses this tradeoff in relation to the worst-instance rate guarantee.

\section{Conclusion and limitations}

Optimizing the deterministic SR checkpoints yields $H_1$-SR, which is optimal
within this class of SR-type algorithms and achieves at least $11/12$ of the optimal Gaussian
$H_1$-normalized rate for every fixed $K$, with the certified fraction tending
to one as $K$ grows; original SR approaches $1/4$. Experiments show substantial
finite-budget gains on many instances but also a clear non-dominance case. The
finite-budget bound extends to independent sub-Gaussian rewards; matching
optimality beyond the Gaussian setting and the early/late elimination tradeoff
remain open.

\subsection*{Acknowledgments}
J. Komiyama was supported by the MBZUAI Start-up fund [BF0121]

\bibliography{references}
\bibliographystyle{arxiv_references}

\clearpage
\appendix
\section{Experimental instances and trajectory plots}
\label{app:experiment-details}

The synthetic configurations follow the benchmark family used by
\citet{komiyama2026anytime}. All ten use $K=40$ independent Gaussian arms
with unit variance. With arms indexed by $i\in\{1,\ldots,40\}$, the means
used in our experiments are
\begin{align*}
\text{Instance 1:}\quad & P_i=1-0.05(i-1),\\
\text{Instance 2:}\quad & P_i=10\left(\frac{i-1}{39}\right)^{0.8},\\
\text{Instance 3:}\quad & P_i=1-\frac{\sqrt{i-1}}{10},\\
\text{Instance 4:}\quad & (P_1,\ldots,P_{40})=(1,0.9,0.9,0.9,0.9,\underbrace{0,\ldots,0}_{35}),\\
\text{Instance 5:}\quad & P_1=\sin\!\left(\frac{39\pi}{80}\right),\quad
P_i=\sin\!\left(\frac{9\pi(40-i-1)}{800}\right),\ 2\le i\le40,\\
\text{Instance 6:}\quad & P_i=0.75\,3^{-i/10},\\
\text{Instance 7:}\quad & (P_1,\ldots,P_{40})=(1,\underbrace{0.8,\ldots,0.8}_{39}),\\
\text{Instance 8:}\quad & (P_1,\ldots,P_{40})=(1,\underbrace{0.8,0.8,0.8}_{3},
\underbrace{0.5,\ldots,0.5}_{6},\underbrace{0.2,\ldots,0.2}_{10},
\underbrace{0,\ldots,0}_{20}),\\
\text{Instance 9:}\quad & (P_1,\ldots,P_{40})=(1,0.8,0.8,\underbrace{0,\ldots,0}_{37}),\\
\text{Instance 10:}\quad & (P_1,\ldots,P_{40})=(1,0.9,0.85,0.8,\underbrace{0,\ldots,0}_{36}).
\end{align*}
For Instances 5 and 8, the displayed definitions state explicitly the
configurations used in our experiments. The two additional problems are
Gaussian instances whose means are derived from MovieLens 1M and Open Bandit,
as described in Section~\ref{sec:experiments}.

\subsection{Trajectory plots}
\begin{figure}[t]
  \centering
  \includegraphics[width=0.96\textwidth]{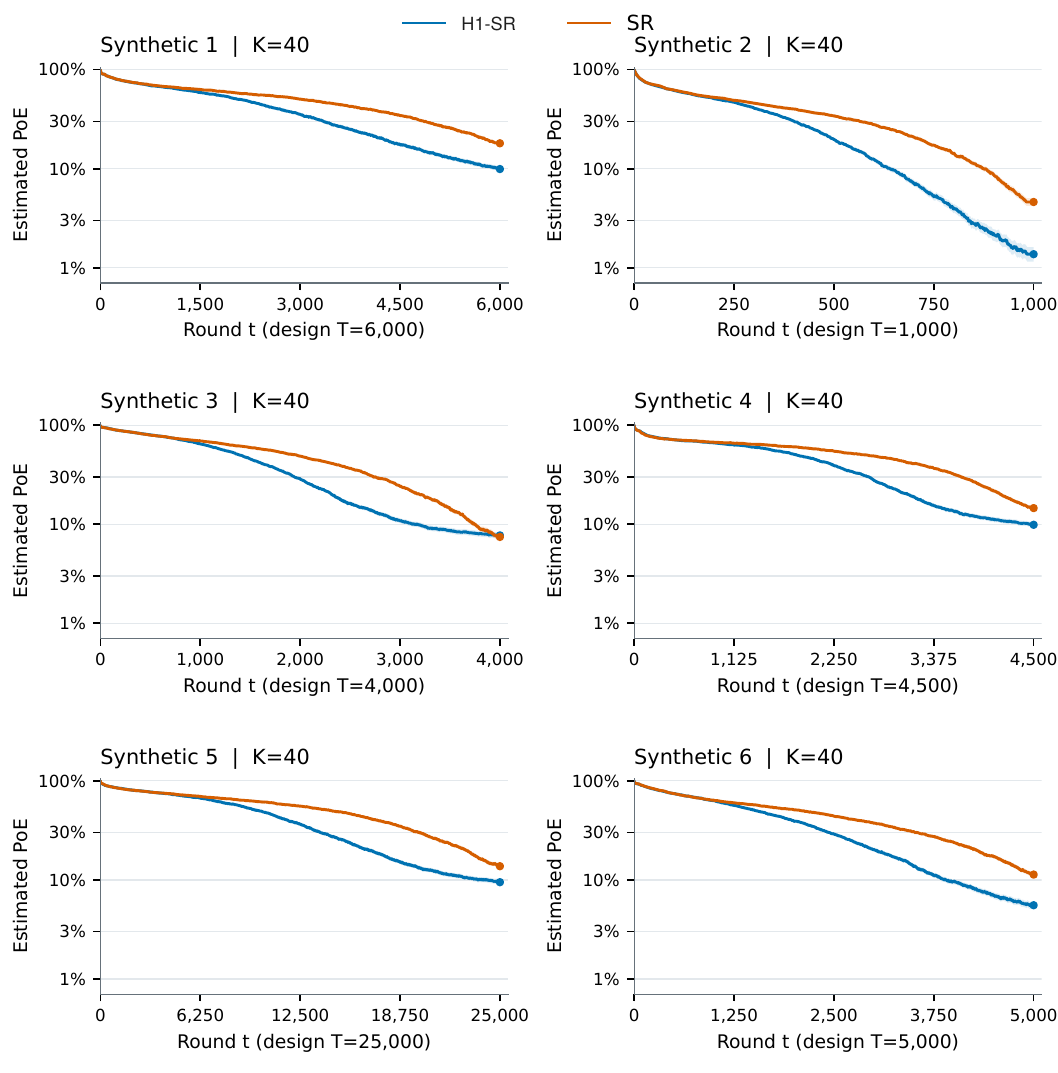}
  \caption{Trajectory comparisons for synthetic instances 1--6. The terminal
  points correspond to the fixed-horizon results in Table~\ref{tab:experiments}.}
  \label{fig:traj1}
\end{figure}

\begin{figure}[t]
  \centering
  \includegraphics[width=0.96\textwidth]{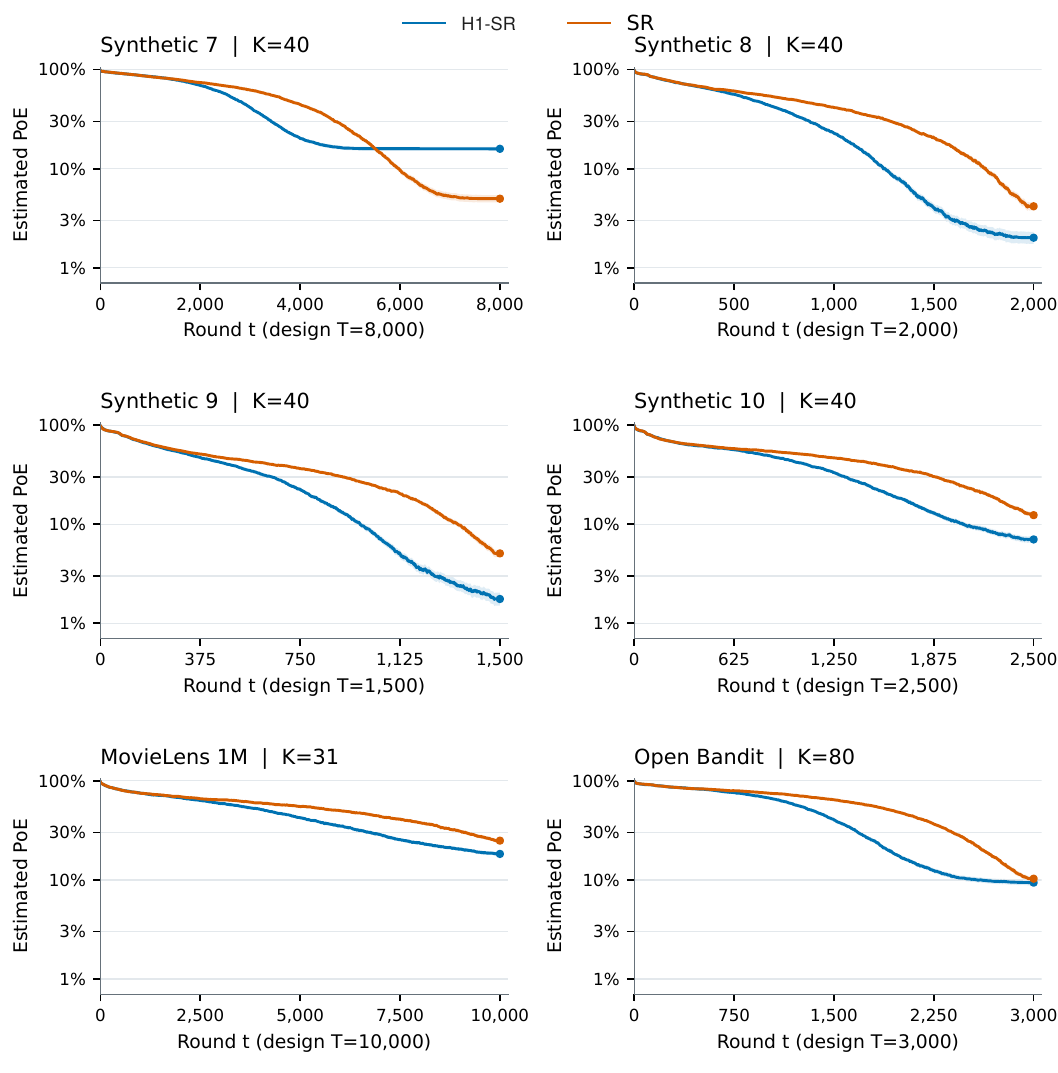}
  \caption{Trajectory comparisons for synthetic instances 7--10, MovieLens 1M,
  and Open Bandit. Synthetic instance 7 illustrates the finite-budget cost of
  allocating less sampling effort before early irreversible eliminations.}
  \label{fig:traj2}
\end{figure}

Figures~\ref{fig:traj1} and~\ref{fig:traj2} report the empirical probability
of error along the trajectories configured for their terminal horizons. The
blue curve is $H_1$-SR and the orange curve is original SR; shading gives
pointwise exact $95\%$ binomial intervals and the vertical axis is logarithmic.
A point at $t<T$ is an intermediate recommendation from the same
$T$-configured run, not a separate run retuned to horizon $t$, so these curves
do not constitute anytime guarantees.

\clearpage

\section{Joint rejection bounds and Gaussian sharpness}\label{app:gaussian}

For the proofs, we use the equivalent construction in which each arm has
an independent stream of potential rewards. Its first $m$ rewards define
the sample mean $\overline Q_{i,m}$, even on paths where the algorithm
does not actually collect all of them. 
At a deterministic count these sample means have their unconditional laws; we never condition those laws on survival. This construction does not change the sequential observation model in the main text.

\begin{lemma}[A fixed-subset rejection bound]\label{lemma:cone}
Suppose independent rewards from every arm are $\sigma^2$-sub-Gaussian.
Fix a deterministic set $A=\{i^\star,i_1,\ldots,i_{s-1}\}$ of $s\ge2$
arms containing the true best arm. Define its true gaps by
$\delta_j=P_{i^\star}-P_{i_j}>0$, and put
$H_A=\sum_{j=1}^{s-1}\delta_j^{-2}$. At a common deterministic count
$m\ge1$, the probability that the best arm has the smallest sample mean
in this set is at most
\begin{equation}\label{eq:app-cone-bound}
 \exp\left\{-\frac{m(s-1)^2}{2s\sigma^2H_A}\right\}.
\end{equation}
\end{lemma}

\begin{proof}
Translate the best mean to zero and write
$\nu=(0,-\delta_1,\ldots,-\delta_{s-1})$. Let $X$ be the vector of
shifted sample means, so that $\E X=\nu$. The joint rejection event is
the cone
\[
 \mathcal C=\{x\in\R^s:x_0\le x_j\text{ for }1\le j<s\}.
\]
Let $x^\circ$ be the Euclidean projection of $\nu$ onto $\mathcal C$, and
set $v=x^\circ-\nu$. The projection inequality gives
$v^\top(x-\nu)\ge\norm{v}^2$ for every $x\in\mathcal C$. Independence
and the sub-Gaussian assumption give, for every $\lambda\in\R$,
\[
 \E\exp\{\lambda v^\top(X-\nu)\}
 \le\exp\left\{\frac{\lambda^2\sigma^2\norm{v}^2}{2m}\right\}.
\]
Applying the Chernoff bound at $\lambda=m/\sigma^2$ therefore implies
\begin{equation}\label{eq:app-projection-bound}
 \Prob(X\in\mathcal C)
 \le \exp\left\{-\frac{m\,\operatorname{dist}(\nu,\mathcal C)^2}
                         {2\sigma^2}\right\}.
\end{equation}

For a proposed first coordinate $-u$, the closest feasible competitor
coordinate is $\max\{-\delta_j,-u\}$. Values $u<0$ cannot improve on
$u=0$. Consequently,
\[
 \operatorname{dist}(\nu,\mathcal C)^2
 =\min_{u\ge0}\left\{u^2+\sum_{j=1}^{s-1}(\delta_j-u)_+^2\right\}.
\]
For any $u\ge0$, put $d_j=(\delta_j-u)_+$ and
$V=u^2+\sum_jd_j^2$. Since $\delta_j\le u+d_j$, Cauchy--Schwarz gives
\begin{align*}
 s-1
 &\le u\sum_j\delta_j^{-1}+\sum_jd_j\delta_j^{-1}\\
 &\le \sqrt V\,
       \sqrt{\left(\sum_j\delta_j^{-1}\right)^2+H_A}
 \le \sqrt{sH_AV}.
\end{align*}
For the second inequality, apply Cauchy--Schwarz to the vectors
\[
 a=(u,d_1,\ldots,d_{s-1}),\qquad
 b=\left(\sum_j\delta_j^{-1},\delta_1^{-1},\ldots,
          \delta_{s-1}^{-1}\right).
\]
Their inner product is $u\sum_j\delta_j^{-1}+\sum_jd_j\delta_j^{-1}$,
while $\norm{a}^2=V$ and
$\norm{b}^2=(\sum_j\delta_j^{-1})^2+H_A$; hence
$a^\top b\le\norm{a}\norm{b}$ gives the stated bound.
The last inequality uses $(\sum_j\delta_j^{-1})^2\le (s-1)H_A$,
again by Cauchy--Schwarz.
Thus $\operatorname{dist}(\nu,\mathcal C)^2\ge(s-1)^2/(sH_A)$.
Together with \eqref{eq:app-projection-bound}, this proves
\eqref{eq:app-cone-bound}, and hence \eqref{eq:joint}.

\end{proof}

The error upper bound alone does not establish an exact exponent.
The following Gaussian lower bound is used in the hard instances of
Appendices~\ref{app:balanced} and~\ref{app:class}.

\begin{lemma}[Exact Gaussian rejection exponent]\label{lemma:gaussian-exponent}
With Gaussian rewards of common variance $\sigma^2$, the rejection
probability for a fixed set satisfies
\[
 \lim_{m\to\infty}-\frac1m\log\Prob(X\in\mathcal C)
 =\frac{\operatorname{dist}(\nu,\mathcal C)^2}{2\sigma^2}.
\]
For equal gaps $\Delta$, this limit is $(s-1)\Delta^2/(2s\sigma^2)$.
\end{lemma}
\begin{proof}
Here $X\sim N(\nu,\sigma^2I_s/m)$. Choose interior points
$y_\varepsilon\in\mathcal C$ converging to
$x^\circ$ and a ball $B(y_\varepsilon,r)$ contained in $\mathcal C$.
Integrating the Gaussian density over that ball yields
\begin{align*}
 \Prob(X\in\mathcal C)
 &\ge \operatorname{vol}(B(y_\varepsilon,r))
       \left(\frac{m}{2\pi\sigma^2}\right)^{s/2}\\[-2pt]
 &\qquad\times
 \exp\left\{-\frac{m}{2\sigma^2}
       \sup_{x\in B(y_\varepsilon,r)}\norm{x-\nu}^2\right\}.
\end{align*}
Take the upper limit of $-m^{-1}\log$ of the probability, then let
$r\downarrow0$ and $\varepsilon\downarrow0$. Combining the result with
\eqref{eq:app-projection-bound} proves
\begin{equation}\label{eq:app-exact-cone-rate}
 \lim_{m\to\infty}-\frac1m\log\Prob(X\in\mathcal C)
 =\frac{\operatorname{dist}(\nu,\mathcal C)^2}{2\sigma^2}.
\end{equation}
For equal gaps, the minimizer has every coordinate equal to
$-(s-1)\Delta/s$, and its squared distance is
$(s-1)\Delta^2/s$. Interior points can be obtained by moving its first
coordinate down by $\varepsilon$ and all other coordinates up by
$\varepsilon$. This proves the claimed equal-gap rate. Ties have
probability zero.
\end{proof}

\section{The balanced clock and its Gaussian worst-instance rate}\label{app:balanced}

\paragraph{Budget feasibility and phase balancing.}
The coefficients $a_s=2s/(s-1)^2$ strictly decrease with $s\ge2$.
For example, their continuous extension has derivative
$-2(s+1)/(s-1)^3<0$. Thus the checkpoints in \eqref{eq:clock} are
nondecreasing across successive rejection phases. Summing each rejected
arm's final cumulative count and the counts of both finalists gives
\begin{align*}
 \text{number of observations used}
 &=2n_{K-1}+\sum_{k=1}^{K-2}n_k\\
 &\le K+\frac{T-K}{D_K}
         \left(2a_2+\sum_{s=3}^Ka_s\right)=T.
\end{align*}
This calculation includes $T=K$. Any remaining observations are taken
after the recommendation is stored and do not affect it.

As in Section~\ref{sec:balance}, $m_s(T)$ is the cumulative
per-arm count at the $s$-arm checkpoint and
$q_s=\lim_{T\to\infty}m_s(T)/T$, whenever the limit exists.
For \HSR, $m_s(T)=n_{K+1-s}$ and $q_s=a_s/D_K$.
For any such limiting schedule the budget constraint is
\begin{equation}\label{eq:app-checkpoint-budget}
 2q_2+\sum_{s=3}^Kq_s\le1.
\end{equation}
If $\gamma=\min_{2\le s\le K}q_s/a_s$, then
$q_s\ge\gamma a_s$ for every $s$. Hence
$1\ge\gamma(2a_2+\sum_{s=3}^Ka_s)=\gamma D_K$.
Equality is attained by $q_s=a_s/D_K$, which satisfies the required
ordering. Equality also forces every coordinate to have this value.
This proves the balancing assertion of Proposition~\ref{prop:balance}.
The statistical upper bound showing
optimality beyond this max--min calculation is proved in
Appendix~\ref{app:class}.

\begin{proof}[Proof of Theorem~\ref{thm:balanced}]
If the best arm is first rejected in phase $k$, its active set has size
$s=K+1-k$. For each possible deterministic active subset $A$ of this
size, the rejection event is contained in the event of
Lemma~\ref{lemma:cone}, evaluated at the deterministic count $n_k$.
The subset complexity satisfies $H_A\le\Hone(\bmP)$, while the rounding
in \eqref{eq:clock} ensures
\[
 \frac{n_k}{a_s}\ge\frac{T-K}{D_K}.
\]
All these subset events share the exponential bound in \eqref{eq:finite},
before multiplication by its prefactor. There are
\[
 \sum_{s=2}^K \binom{K-1}{s-1}=2^{K-1}-1
\]
subsets containing the best arm and at least one competitor. A union
bound, and the trivial bound by one, prove the finite-budget statement.
Taking the horizon limit at fixed $K$ and a fixed instance gives only
the lower bound
\[
 \liminf_{T\to\infty}\frac1T
 \log\frac1{\PoE_{\HSR}(\bmP)}
 \ge\frac1{\sigma^2D_K\Hone(\bmP)}.
\]
For the Gaussian model defining the normalized rate, it follows that
\[
 R_{\HSR}\ge\frac1{\sigma^2D_K}.
\]

\end{proof}

\section{Optimality within the class of SR-type algorithms}\label{app:class}

The following proof goes as following:
For each $s$, we construct an instance that is difficult at the
checkpoint with $s$ arms remaining, and then take the worst case over $s$.

\begin{proof}[Proof of Theorem~\ref{thm:class}]
Work in the Gaussian model and consider an algorithm $\mathcal A$ in
the class of SR-type algorithms. In particular, each rejection
removes an empirical worst arm; only its deterministic common checkpoints
$m_K(T)\le\cdots\le m_2(T)$ are allowed to change. By compactness, along
a horizon subsequence all fractions $m_s(T)/T$ converge to limits $q_s$.
They satisfy $0\le q_K\le\cdots\le q_2$ and
\eqref{eq:app-checkpoint-budget}. If $R_{\mathcal A}=0$, the result
is immediate. Otherwise fix any finite $0<r<R_{\mathcal A}$.

A positive rate requires $q_K>0$. Indeed, on a fixed equal-gap
Gaussian instance the first-checkpoint rejection probability is at least
$c_0e^{-c_1m_K(T)}$ for constants $c_0,c_1>0$. This follows from the
Gaussian ball bound above, uniformly over integer counts $m_K(T)\ge1$.
If $q_K=0$, then $m_K(T)=o(T)$ along the subsequence, so the
full-sequence lower error exponent is zero, contradicting
$R_{\mathcal A}>0$. This boundary case is needed before choosing the
far-arm gap below.

Fix $s\in\{2,\ldots,K\}$ and $\Delta>0$. For a fixed $M>\Delta$,
consider the instance with one best mean~0, $s-1$ near means
$-\Delta$, and $K-s$ far means $-M$. Let $E_T$ be the unconditioned
event that all near competitors have sample means larger than the best
at count $m_s(T)$. Lemma~\ref{lemma:gaussian-exponent} gives, along the selected
subsequence,
\begin{equation}\label{eq:app-class-near-event}
 \Prob(E_T)=\exp\{-TI_s+o(T)\},
 \qquad
 I_s=\frac{q_s(s-1)\Delta^2}{2s\sigma^2}.
\end{equation}

Let $S_T$ denote the event that all $s$ near arms, including the best,
survive to the $s$-arm checkpoint. If $S_T$ fails, some near arm is
rejected at an earlier checkpoint while a far arm remains active.
Because the rejected arm is empirically worst, at that checkpoint a far
sample mean is at least as large as a near sample mean. Every relevant
true gap is at least $M-\Delta$, and every checkpoint count is at least
$m_K(T)$. The Gaussian two-arm tail and a finite union over arms and
earlier checkpoints yield a constant $L_K<\infty$ such that
\begin{equation}\label{eq:app-class-early-event}
 \Prob(S_T^c)
 \le L_K\exp\left\{-\frac{m_K(T)(M-\Delta)^2}{4\sigma^2}\right\}.
\end{equation}
For $s=K$, the event $S_T$ is certain and this bound is unnecessary.
Choose $M$ large enough that
$q_K(M-\Delta)^2/(4\sigma^2)>I_s$. On $E_T\cap S_T$ the best arm is
rejected at the $s$-arm checkpoint. Thus, without any independence
assumption between these events,
\[
 \PoE_{\mathcal A}(\bmP)
 \ge\Prob(E_T)-\Prob(S_T^c)
 =\exp\{-TI_s+o(T)\}.
\]
The full-sequence lower exponent is no larger than the corresponding
upper exponent along this subsequence. Since
$\Hone(\bmP)=(s-1)/\Delta^2+(K-s)/M^2$, we obtain
\[
 r\le\left(\frac{s-1}{\Delta^2}+\frac{K-s}{M^2}\right)I_s.
\]
Here the horizon limit has been taken with $M$ fixed. Now let
$M\to\infty$, giving
\[
 r\le\frac{q_s(s-1)^2}{2s\sigma^2}
     =\frac{q_s}{\sigma^2a_s}.
\]
This holds for every $s\in\{2,\ldots,K\}$, so
\[
 r\le\min_{2\le s\le K}\frac{q_s}{\sigma^2a_s}.
\]
Thus the smallest of these bounds over all checkpoints limits the rate.
Equivalently, $q_s\ge\sigma^2r a_s$ for every $s$, and
\eqref{eq:app-checkpoint-budget} implies $1\ge\sigma^2rD_K$.
Since this holds for every finite $0<r<R_{\mathcal A}$, it gives
$R_{\mathcal A}\le1/(\sigma^2D_K)$. Theorem~\ref{thm:balanced}
and Appendix~\ref{app:balanced} show that \HSR{} attains this value.
The proof
does not assume that a schedule's normalized checkpoints converge along
the full horizon sequence.
\end{proof}

\section{A converse for all adaptive algorithms}\label{app:converse}

The one-shot \textit{game-optimal} bound $R^{\mathrm{go}}$ in
\citet{komiyama2022minimax,komiyama2026anytime} is an upper bound on
attainable normalized rates for any algorithm. Our argument uses essentially the same change-of-measure
principle. For Gaussian rewards with common variance $\sigma^2$ and
the complexity $H_1$ used here, our construction gives $B_K$ such that
\[
 R_K^\star\le R^{\mathrm{go}}\le\frac1{\sigma^2B_K}.
\]
Thus $1/(\sigma^2B_K)$ is an explicit upper bound on the one-shot game
value, but not necessarily its exact value, and it is a useful simple formula.
We first prove the allocation statement needed to impose several
constraints on one common vector. An algorithm may choose which arm to
sample using past observations, internal randomness, and the known budget
$T$. The true arm means are unknown and are not inputs to the algorithm.
Throughout this section rewards are Gaussian with the
common variance $\sigma^2$. We use $\boldsymbol\theta$ for the true mean vector of a reference
instance whose unique best arm is arm~1, and $\bmP$ for the true
mean vector of an alternative instance with a different best arm.

The following allocation statement follows directly from the definition
of $R^{\mathrm{go}}$ in
\citet{komiyama2022minimax,komiyama2026anytime}
and holds for every finite $0<r<R^{\mathrm{go}}$. We derive it here for completeness.
\begin{lemma}[One allocation for every fixed alternative]
\label{lemma:app-allocation}
Fix a finite $r$ with $0<r<R^{\mathrm{go}}$. For any fixed unique-best
instance $\boldsymbol\theta$ with best arm~1, there exists
$u(\boldsymbol\theta)$ in the probability simplex such that, for every
fixed unique-best instance $\bmP$ whose best arm is not~1,
\begin{equation}\label{eq:app-allocation}
 r\le\frac{\Hone(\bmP)}{2\sigma^2}
          \sum_{i=1}^K u_i(\boldsymbol\theta)(\theta_i-P_i)^2.
\end{equation}
\end{lemma}

\begin{proof}
Since $r<R^{\mathrm{go}}$, the definition of the one-shot game provides
an allocation map $w(\cdot)$ into the probability simplex such that
\[
 r\le \Hone(\bmP)\sum_{i=1}^K w_i(\boldsymbol\theta)
       \KL\!\left(N(\theta_i,\sigma^2)\,\|\,N(P_i,\sigma^2)\right)
\]
for every unique-best reference instance $\boldsymbol\theta$ and every
unique-best alternative $\bmP$ with a different best arm.
For Gaussian distributions with common variance, the KL divergence is
$(\theta_i-P_i)^2/(2\sigma^2)$. Setting
$u(\boldsymbol\theta)=w(\boldsymbol\theta)$ proves
\eqref{eq:app-allocation}. The same allocation map works for all these
pairs; in particular, $u(\boldsymbol\theta)$ does not depend on the
alternative $\bmP$. The strict inequality $r<R^{\mathrm{go}}$ avoids
assuming that the supremum defining the game value is attained.
\end{proof}

We now use this allocation statement to prove
$R^{\mathrm{go}}\le1/(\sigma^2B_K)$.
We first derive constraints on a common allocation vector using Gaussian
instances with gaps at different scales, then show that these constraints
imply $1\ge\sigma^2rB_K$ for every finite $0<r<R^{\mathrm{go}}$.
For a nonnegative vector $u$ and $2\le s\le K$, define
\begin{equation}\label{eq:app-prefix-information}
 \Psi_s(u)=\inf_{\substack{z_i>0,\ i<s\\t\in\R}}
 \left(\sum_{i<s}z_i^{-2}\right)
 \left[u_st^2+\sum_{i<s}u_i(z_i-t)^2\right].
\end{equation}

\begin{lemma}[Simultaneous prefix information constraints]
\label{lemma:app-prefixes}
For every finite $r$ with $0<r<R^{\mathrm{go}}$, there exists one
probability vector $u$ satisfying
\begin{equation}\label{eq:app-prefix-constraints}
 \Psi_s(u)\ge2\sigma^2r,\qquad 2\le s\le K.
\end{equation}
\end{lemma}

\begin{proof}
For $0<\varepsilon<1$, choose the strictly ranked reference instance
\[
 \theta_1^\varepsilon=0,
 \qquad
 \theta_j^\varepsilon=-\sum_{\ell=2}^j
                    \varepsilon^{2(K-\ell+1)},\quad j\ge2.
\]
Apply Lemma~\ref{lemma:app-allocation} for each fixed
$\boldsymbol\theta^\varepsilon$. Along a sequence
$\varepsilon\downarrow0$, extract a
limit $u$ of the resulting allocation vectors. For each $s\in\{2,\ldots,K\}$,
we call the arms $\{1,\ldots,s\}$ the \emph{prefix} and the remaining
arms $\{s+1,\ldots,K\}$ the \emph{suffix}, which is empty when $s=K$.
Put $L_s=\varepsilon^{2(K-s)+1}$. Then
\[
 \theta_i^\varepsilon/L_s\to0\quad(i\le s),
 \qquad
 \theta_i^\varepsilon/L_s\to-\infty\quad(i>s).
\]
For fixed positive $z_1,\ldots,z_{s-1}$ and $t\ge0$, define
\[
 P_s^\varepsilon=L_st,
 \qquad
 P_i^\varepsilon=L_s(t-z_i)\ (i<s),
 \qquad
 P_i^\varepsilon=\theta_i^\varepsilon\ (i>s).
\]
For sufficiently small $\varepsilon$, arm $s$ is the unique best arm,
and the suffix contributes asymptotically nothing to the scaled
complexity:
\[
 L_s^2\Hone(\bmP^\varepsilon)\longrightarrow
 \sum_{i<s}z_i^{-2}.
\]
The suffix means are unchanged, so its contribution to the quadratic
information cost is exactly zero. Applying \eqref{eq:app-allocation}
and then taking $\varepsilon\downarrow0$ yields
\[
 2\sigma^2r\le
 \left(\sum_{i<s}z_i^{-2}\right)
 \left[u_st^2+\sum_{i<s}u_i(z_i-t)^2\right].
\]
For fixed positive $z$, the minimizing translation is
$t=(\sum_{i<s}u_i z_i)/(\sum_{i\le s}u_i)\ge0$ whenever the denominator
is positive. If all prefix coordinates vanish, the expression is zero,
contradicting $r>0$. Thus the restriction $t\ge0$ loses no minimizing
value, and taking the infimum proves \eqref{eq:app-prefix-constraints}.

The allocation map is fixed before taking $\varepsilon\downarrow0$.
The convergent subsequence of allocation vectors is common to all the
finitely many prefixes and all fixed choices of their parameters.
No horizon limit is needed in this argument.
\end{proof}

\begin{proof}[Proof of Theorem~\ref{thm:converse}]
If $R^{\mathrm{go}}=0$, the claimed bound is immediate. Otherwise fix any
finite $0<r<R^{\mathrm{go}}$ and use the vector in
Lemma~\ref{lemma:app-prefixes}. Write $c=\sigma^2r$. At $s=2$, direct
minimization gives
\[
 \Psi_2(u)=\frac{u_1u_2}{u_1+u_2}\le\frac{u_1+u_2}{4}.
\]
The denominator is positive by the prefix constraint. Hence
$u_1+u_2\ge8c$. For $K=2$, this already gives $1\ge cB_2$ with $B_2=8$.
Now suppose $K\ge3$.
For $s\ge3$, choose $z_i=1$ in
\eqref{eq:app-prefix-information} and minimize over $t$. With
$C_s=\sum_{i<s}u_i$, this gives
\[
 \Psi_s(u)\le(s-1)\frac{C_su_s}{C_s+u_s}
             \le(s-1)u_s.
\]
Thus $u_s\ge2c/(s-1)$. To retain the coupling between the first three
arms, put $C=u_1+u_2\ge8c$. The same inequality at $s=3$, before
discarding its denominator, gives
\[
 \frac{Cu_3}{C+u_3}\ge c,
 \qquad
 u_3\ge\frac{cC}{C-c}.
\]
The function $C\mapsto C^2/(C-c)$ is increasing for $C\ge8c$, since its
derivative is $C(C-2c)/(C-c)^2>0$. Consequently,
\[
 u_1+u_2+u_3\ge\frac{C^2}{C-c}\ge\frac{64c}{7}.
\]
Combining this with the individual bounds for $s\ge4$ yields
\[
 1=\sum_i u_i
 \ge c\left(\frac{64}{7}+2\sum_{s=4}^K\frac1{s-1}\right)
 =c\left(\frac{43}{7}+2h_{K-1}\right)=\sigma^2r B_K.
\]
An empty sum is zero, so this includes $K=3$. Because the inequality
holds for every finite $0<r<R^{\mathrm{go}}$, it implies
$R^{\mathrm{go}}\le1/(\sigma^2B_K)$, including ruling out an infinite
game value. Combining this with the bound
$R_K^\star\le R^{\mathrm{go}}$ from
\citet{komiyama2022minimax,komiyama2026anytime} proves the theorem.
\end{proof}

\section{The exact exponent of original Successive Rejects}\label{app:sr}

The exact error exponent of original SR is established in prior work by \cite{DBLP:conf/icml/WangAP24} and was rediscovered by \cite{komiyama2026anytime}. 
Here, we will re-derive it for completeness.

\begin{proof}[Proof of Theorem~\ref{thm:sr}]
We first work with Gaussian rewards and prove the finite-budget bound
for $T>K$. At phase $k$, put
$s=K+1-k$. If the true best is rejected, at least one true rank
$r\in\{s,\ldots,K\}$ remains active and has a sample mean no smaller
than the best. Indeed, among $s$ active arms there must be one of those
$K-s+1$ ranks, since only $s-1$ arms have better true ranks.
For a deterministic common count $m$,
\[
 \Prob\{\overline Q_{(r),m}\ge\overline Q_{i^\star,m}\}
 =\Phi\left(-\Delta_{(r)}\sqrt{\frac{m}{2\sigma^2}}\right)
 \le\frac12\exp\left\{-\frac{m\Delta_{(r)}^2}{4\sigma^2}\right\}.
\]
Here $\Phi$ is the standard normal distribution function; the displayed
tail inequality follows, for example, by applying the Gaussian density
shift to its integral over the positive half-line.
The SR checkpoint satisfies
$n_k^{\mathrm{SR}}\ge(T-K)/(s\barlog(K))$. Also
$\Hone(\bmP)\ge(s-1)/\Delta_{(s)}^2$ and
$\Delta_{(r)}\ge\Delta_{(s)}$. Therefore
\[
 n_k^{\mathrm{SR}}\Delta_{(r)}^2
 \ge\frac{T-K}{\barlog(K)\Hone(\bmP)}\frac{s-1}{s}
 \ge\frac{T-K}{2\barlog(K)\Hone(\bmP)}.
\]
At phase $k$, there are $k$ possible ranks
$r\in\{K+1-k,\ldots,K\}$ to consider. Summing over all phases
gives $\sum_{k=1}^{K-1}k=K(K-1)/2$ events in the union bound.
The union bound proves the asserted prefactor $K(K-1)/4$ and exponent, with a
further minimum with one if necessary. The explicit one-sample-per-arm
convention at $T=K$ removes the zero-count endpoint of the original
checkpoint formula and does not affect any asymptotic claim.

For independent $\sigma^2$-sub-Gaussian rewards, the difference of the
two sample means instead has variance proxy $2\sigma^2/m$. The Chernoff
bound gives
\[
 \Prob\{\overline Q_{(r),m}\ge\overline Q_{i^\star,m}\}
 \le\exp\left\{-\frac{m\Delta_{(r)}^2}{4\sigma^2}\right\}.
\]
The same union bound over phases and ranks therefore proves the same finite-budget
exponential term with prefactor $K(K-1)/2$, rather than $K(K-1)/4$.
The additional factor $1/2$ per pairwise event above is Gaussian-specific.

To prove Gaussian sharpness, first let $K\ge3$ and consider the fixed instance
\[
 \bmP^{(M)}=(0,-1,-M,\ldots,-M),\qquad M>1.
\]
Write $m=n_{K-1}^{\mathrm{SR}}$ and define the unconditioned final
reversal event $E_T=\{\overline Q_{2,m}\ge\overline Q_{1,m}\}$.
Since $m/T\to1/(2\barlog(K))$, the normal-tail exponent gives
\begin{equation}\label{eq:app-sr-reversal}
 \Prob(E_T)=\exp\left\{-\frac{T}{8\sigma^2\barlog(K)}+o(T)\right\}.
\end{equation}
Let $S_T$ be the event that arms 1 and 2 both reach the final phase.
If either is removed earlier, a far arm must then have a sample mean
at least as large. The smallest relevant gap is $M-1$ and the smallest
checkpoint fraction tends to $1/(K\barlog(K))$. A finite union bound
consequently gives
\begin{equation}\label{eq:app-sr-early-loss}
 \Prob(S_T^c)
 \le\exp\left\{-\frac{T(M-1)^2}{4\sigma^2K\barlog(K)}+o(T)\right\}.
\end{equation}
Choose fixed $M$ such that $(M-1)^2>K/2$. The exponent in
\eqref{eq:app-sr-early-loss} is strictly larger than the exponent in
\eqref{eq:app-sr-reversal}. On $S_T\cap E_T$, SR rejects arm~1 in
its final phase. Hence
\[
 \PoE_{\mathrm{SR}}(\bmP^{(M)})
 \ge\Prob(E_T)-\Prob(S_T^c)
 =\exp\left\{-\frac{T}{8\sigma^2\barlog(K)}+o(T)\right\}.
\]
Since $\Hone(\bmP^{(M)})=1+(K-2)/M^2$, first taking $T\to\infty$
for this fixed instance and then $M\to\infty$ proves
\[
 R_{\mathrm{SR}}\le\frac1{8\sigma^2\barlog(K)}.
\]
For $K=2$, the same upper bound follows directly from the final
two-arm Gaussian comparison. The finite-budget guarantee supplies the
reverse inequality for every $K$, proving exact equality.
\end{proof}

\section{Uniform factors and the large-arm limit}\label{app:factors}

\begin{proof}[Proof of Corollary~\ref{cor:factor}]
Theorems~\ref{thm:balanced} and~\ref{thm:converse} immediately imply
\[
 \frac{R_{\HSR}}{R_K^\star}\ge\frac{B_K}{D_K}.
\]
For $K=2$, $B_2=D_2=8$ and the algorithm is optimal.
We use $11/12$ as a convenient rational lower bound, rather than as
the exact minimum of $B_K/D_K$ that is a bit involved. 
Since $D_K>0$, it suffices to show that $12B_K-11D_K\ge0$ for every $K$.
For $K\ge3$, put $m=K-1$ and define
\[
 G_m=12B_K-11D_K
     =\frac{208}{7}+2h_m-22h_m^{(2)}.
\]
For $m\ge3$, its successive differences are
\[
 G_m-G_{m-1}=\frac{2(m-11)}{m^2}.
\]
Thus, over the integers $m\ge2$, the minimum is attained at $m=10$ and
$m=11$. Exact arithmetic gives
\[
 G_{10}=G_{11}=\frac{187633}{127008}>0.
\]
It follows that $B_K/D_K>11/12$ for every $K\ge3$, proving the uniform
factor. 

The exact SR rate and the all-algorithm sandwich give
\[
 \frac{B_K}{8\barlog(K)}
 \le\frac{R_{\mathrm{SR}}}{R_K^\star}
 \le\frac{D_K}{8\barlog(K)}.
\]
Moreover,
\[
 \frac{B_K}{8\barlog(K)}
 \ge\frac14+\frac{7-2/K}{8\barlog(K)}>\frac14,
\]
because $B_K\ge6+2h_{K-1}$ and
$\barlog(K)=h_{K-1}+1/K-1/2$.
Since $h_{K-1}=\log K+O(1)$ and $h_{K-1}^{(2)}=O(1)$, both
$B_K$ and $D_K$ equal $2\log K+O(1)$, whereas
$8\barlog(K)=8\log K+O(1)$. More precisely, for $K\ge3$,
\[
 D_K-B_K=2h_{K-1}^{(2)}-\frac{15}{7}=O(1).
\]
Using $h_{K-1}=\log K+\gamma+O(K^{-1})$ and
$h_{K-1}^{(2)}=\pi^2/6+O(K^{-1})$, where $\gamma$ is Euler's
constant, also gives the first-order expansion
\[
 \frac{B_K}{D_K}
 =1-\frac{\pi^2/6-15/14}{\log K}
   +O\!\left((\log K)^{-2}\right).
\]
Since the achieved rate cannot exceed the all-algorithm optimum,
\[
 0\le1-\frac{R_{\HSR}}{R_K^\star}
 \le\frac{D_K-B_K}{D_K}=O\!\left(\frac1{\log K}\right).
\]
The preceding bounds also give
\[
 \frac{R_{\HSR}}{R_K^\star}\longrightarrow1,
 \qquad
 \frac{R_{\mathrm{SR}}}{R_K^\star}\longrightarrow\frac14.
\]
The exact improvement of the balanced rate over SR is
$8\barlog(K)/D_K\to4$. These are limits of fixed-$K$ asymptotic
error exponents: the horizon limit is taken first. In particular, no
claim about discarding a $K$-dependent finite-budget prefactor when
$K$ grows with $T$ is needed.
\end{proof}

\section{Non-dominance of $\HSR$ over original SR}\label{app:non-dominance}

At a fixed budget $T$, we say that an algorithm $\mathcal A$ dominates
another algorithm $\mathcal A'$ if
\[
 \PoE_{\mathcal A}(\bmP)\le\PoE_{\mathcal A'}(\bmP)
 \quad\text{for every unique-best instance }\bmP,
\]
with strict inequality for at least one instance. Both probabilities
are evaluated at the same budget and under the same true reward
distributions. 

The optimality of $\HSR$ concerns the worst-instance normalized
asymptotic rate $R_{\mathcal A}$. Maximizing this criterion can involve
tradeoffs between performance on different instances $\bmP$ and
$\bmP'$. It does not imply dominance over original SR at a fixed budget.

Synthetic instance 7 illustrates this distinction. It has one best arm
with mean $1$ and $39$ competitors with mean $0.8$, all with unit-variance
Gaussian rewards. In this case, the first elimination matters most to the probability of correctly identifying the best arm.
At $T=8000$, $\HSR$ takes $27$ observations per arm
before the first elimination, compared with $53$ for original SR.
In other words, $\HSR$ allocates fewer samples
before early eliminations. As a result, SR outperforms \HSR{} in this instance.

This provides empirical evidence against finite-budget dominance by
$\HSR$, despite its better worst-instance normalized asymptotic rate. 
Such tradeoffs across underlying distributions are common in statistical
decision theory \citep{Jenninson1982}: a procedure can perform better on some distributions and worse on others, so neither procedure dominates the other.

This differs from fixed-confidence best-arm identification, where
algorithms such as Track-and-Stop can attain the instance-dependent
lower bound on expected sample complexity asymptotically as
$\delta\to0$ \citep{garivier2016optimal}.
Such algorithms are asymptotically optimal on each instance,
so attaining the optimal leading-order sample complexity on one
instance does not require sacrificing it on another.
In contrast, our optimality guarantee concerns the worst-instance
normalized asymptotic rate. Within the class of SR-type algorithms,
improving performance on some instances can come at the cost of
worse performance on others.

\section{The inverse-gap Complexity Measures}\label{app:normalization}

Our complexity is the sum over suboptimal arms only:
\[
 \Hone(\bmP)=\sum_{i\ne i^\star}\Delta_i^{-2}.
\]
In the original SR paper, \citet{audibert2010best} assign the best arm
the artificial gap $\Delta_{(2)}$ and include its inverse square in
their complexity. Writing their complexity as $H_1^{\mathrm{SR}}$, we have
\[
 H_1^{\mathrm{SR}}=\Hone+\Delta_{(2)}^{-2},
 \qquad
 \frac{K}{K-1}\Hone\le H_1^{\mathrm{SR}}\le2\Hone.
\]
Indeed, $\Delta_{(2)}^{-2}$ is the largest of the $K-1$ terms in
$\Hone$, so it lies between $\Hone/(K-1)$ and $\Hone$.

\end{document}